\documentclass[10pt,twocolumn,letterpaper]{article}

\usepackage{cvpr}
\usepackage{times}
\usepackage{epsfig}
\usepackage{graphicx}
\usepackage{amsmath}
\usepackage{amssymb}

\usepackage{multirow}
\usepackage{tabularx}
\usepackage{booktabs}
\usepackage{amssymb}
\usepackage{pifont}

\usepackage{setspace}

\usepackage[dvipsnames]{xcolor}

\newcommand{\floor}[1]{\lfloor #1 \rfloor}
\newcommand{\ceil}[1]{\lceil #1 \rceil}

\newcommand{\cmark}{\ding{51}}%
\newcommand{\xmark}{\ding{55}}%

\newcommand{\G}{\mathcal{G}}
\newcommand{\F}{\mathcal{F}}
\newcommand{\E}{\mathcal{E}}

\renewcommand{\H}{\mathcal{H}}

\usepackage{algorithm} 
\usepackage{algorithmic}

\cvprfinalcopy 

\def\cvprPaperID{576} 

\begin{document}

\title{G-TAD: Sub-Graph Localization for Temporal Action Detection \\ \small\url{https://www.deepgcns.org/app/g-tad }}

\author{Mengmeng Xu,\quad Chen Zhao,\quad David S. Rojas,\quad Ali         Thabet,\quad Bernard Ghanem\\
		King Abdullah University of Science and Technology (KAUST), Saudi Arabia\\
		{\tt\footnotesize \{mengmeng.xu, chen.zhao, davidsantiago.blanco, ali.thabet, bernard.ghanem\}@kaust.edu.sa}}

\maketitle

\begin{abstract}
Temporal action detection is a fundamental yet challenging task in video understanding. Video context is a critical cue to effectively detect actions, but current works mainly focus on temporal context, while neglecting semantic context as well as other important context properties. In this work, we propose a graph convolutional network (GCN) model to adaptively incorporate  multi-level semantic context into video features and cast temporal action detection as a sub-graph localization problem. Specifically, we formulate video snippets as graph nodes, snippet-snippet correlations as edges, and actions associated with context as target sub-graphs. With graph convolution as the basic operation, we design a GCN block called GCNeXt, which learns the features of each node by aggregating its context and dynamically updates the edges in the graph. To localize each sub-graph, we also design an SGAlign layer to embed each sub-graph into the Euclidean space. Extensive experiments show that G-TAD is capable of finding effective video context without extra supervision and achieves state-of-the-art performance on two detection benchmarks. On ActivityNet-1.3, it obtains an average mAP of $34.09\%$; on THUMOS14, it reaches $51.6\%$ at IoU@0.5 when combined with a proposal processing method. G-TAD code is publicly available at \href{https://github.com/frostinassiky/gtad}{https://github.com/frostinassiky/gtad}.


\end{abstract}

\section{Introduction}
Video understanding has gained much attention from both academia and industry over recent years, given the rapid growth of videos published in online platforms. Temporal action detection is one of the interesting but challenging tasks in this area.
It involves detecting the start and end frames of action instances, as well as predicting their class labels. This is onerous especially in long untrimmed videos. 

\begin{figure}[t]
    \centering
    \includegraphics[trim={2cm 4cm 3cm 2.5cm},width=9cm,clip]{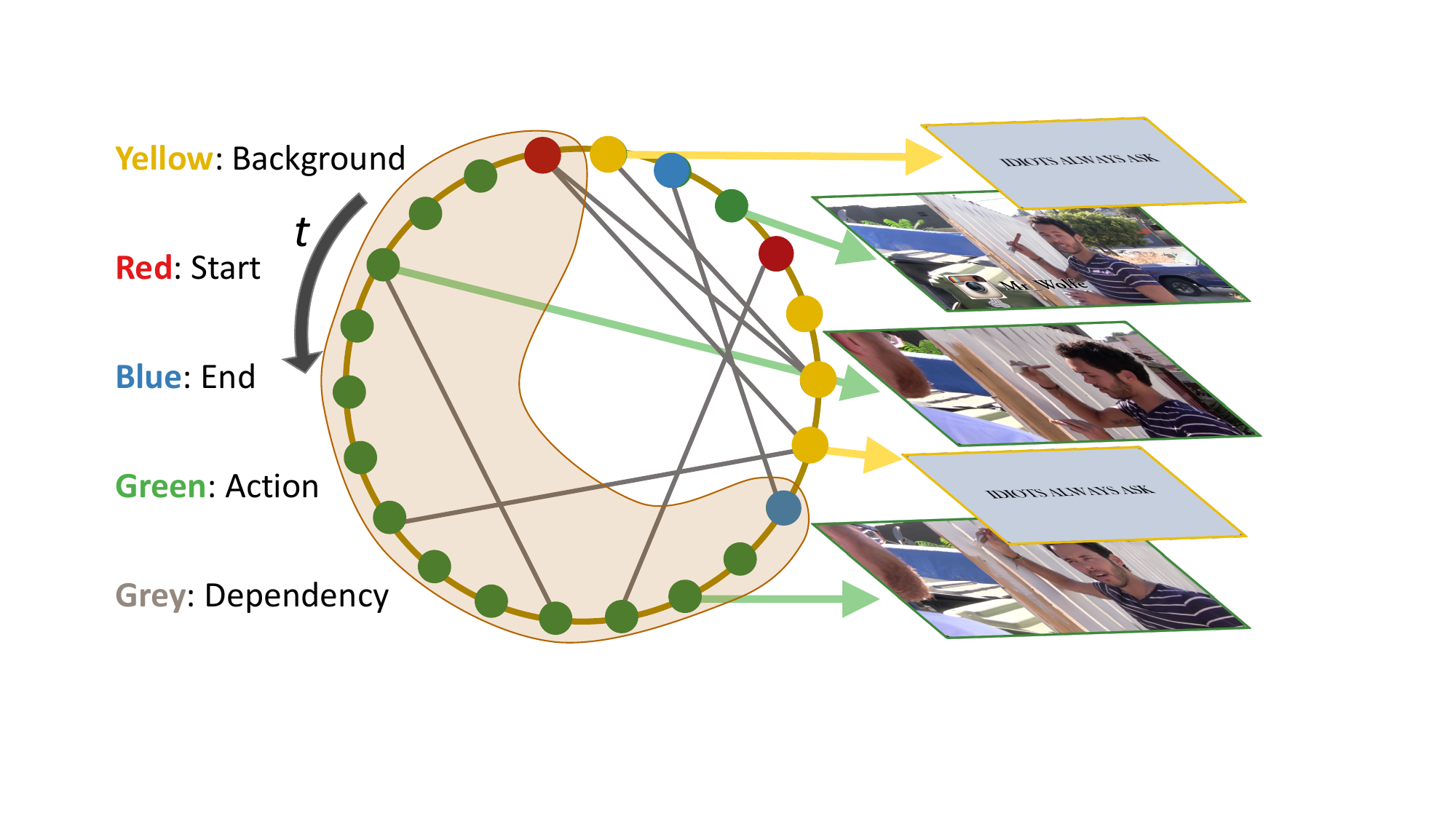}
    \caption{\textbf{Graph formulation of a video.} Nodes: video snippets (a video snippet is defined as consecutive frames within a short time period). Edges:  snippet-snippet correlations. Sub-graphs: actions associated with context. There are 4 types of nodes: action, start, end, and background, shown as colored dots. There are 2 types of edges: (1) temporal edges, which are pre-defined according to the snippets' temporal order; (2) semantic edges, which are learned from node features.} 
    \label{fig:intro}
\end{figure}


Video context is an important cue to effectively detect actions. Here, we refer to context as  
 frames that do not belong to the target action but carry its valuable indicative information. 
Using video context to infer potential actions is natural for human beings. In fact, empirical evidence shows that humans can reliably guess or predict the occurrence of a certain type of action by only looking at short video snippets where the action does not happen~\cite{detad, Alwassel2017ActionSS}. 
Therefore, incorporating context into temporal action detection has become an important strategy to boost detection accuracy in the recent literature \cite{ dai2017temporal, gao2017turn, chao2018rethinking, Long2019GaussianTA, TSN2016ECCV, zhao2017temporal, lin2018bsn}. Researchers have proposed various ways to take advantage of video context, such as extending temporal action boundaries by a pre-defined ratio~\cite{dai2017temporal, gao2017turn, TSN2016ECCV, zhao2017temporal, lin2018bsn}, using dilated convolution to encode context into features~\cite{chao2018rethinking}, and aggregating context features implicitly by way of a Gaussian curve~\cite{Long2019GaussianTA}. All these methods only utilize \textit{temporal context}, which precedes or follows an action instance in its immediate temporal neighborhood. However, real-world videos vary dramatically in temporal extent, action content, and even editing preferences. \textit{Temporal context} does not fully exploit the rich merits of video context, and it may even impair detection accuracy if not properly designed for underlying videos. 

So, what properties characterize desirable video context for the purpose of accurate action detection? First, context should be semantically correlated to the target action other than merely temporally located in its vicinity. Imagine the case where we manually stitch an action clip into some irrelevant frames, the abrupt scene change surrounding the action would definitely not benefit the action detection. On the other hand, snippets located at a distance from an action but containing similar semantic content might provide indicative hints for detecting the action. Second, context should be content-adaptive rather than manually pre-defined. Considering the vast variation of videos, context that helps to detect different action instances could be different in lengths and locations based on the video content. Third, context should be based on multiple semantic levels, since using only one form/level of context is unlikely to generalize well.

%

We endow video context with all the above properties by casting action detection as a sub-graph localization problem based on a graph convolutional network (GCN)~\cite{kipf2016semi}. We represent each video sequence as a graph, each snippet as a node, each snippet-snippet correlation as an edge, and target action associated with context as sub-graph, as shown in Fig.~\ref{fig:intro}. The context of a snippet is considered to be all snippets connected to it by edges in a video graph. We define two types of edges --- temporal edges and semantic edges, corresponding to temporal context and semantic context, respectively. Temporal edges exist between each pair of adjacent snippets, whereas semantic edges are dynamically learned from the video features at each layer. Hence, multi-level context of each snippet is gradually aggregated into the features of the snippet throughout the entire GCN. 


The pipeline of our proposed Graph-Temporal Action Detection method, dubbed  G-TAD, is analogous to faster R-CNN~\cite{girshick2015fast, ren2015faster} in object detection. There are two critical designs in G-TAD. First, the GCN-based feature extraction block GCNext, which is inspired by ResNeXt~\cite{xie2017aggregated},  generates context-enriched features. It corresponds to the CNN blocks of the backbone network in faster R-CNN. Second, to mimic region of interest (RoI) alignment~\cite{he2017mask}, we design a sub-graph of interest alignment layer SGAlign to generate a fixed-size representation for each sub-graph and embed all sub-graphs into the same Euclidean space. Finally, we apply a classifier on the features of each sub-graph to obtain detection.  We summarize our contributions as follows.

\noindent
\textbf{(1)}  We present a novel GCN-based video model to fully exploit video context for effective temporal action detection. Using this video GCN representation, we are able to adaptively incorporate multi-level semantic context into the features of each snippet.

\noindent
\textbf{(2)} We propose G-TAD, a new sub-graph detection framework to localize actions in video graphs. G-TAD includes two main modules: GCNeXt and SGAlign. GCNeXt performs graph convolutions on video graphs, leveraging both temporal and semantic context. SGAlign re-arranges sub-graph features in an embedded space suitable for detection.

\noindent
\textbf{(3)} G-TAD achieves state-of-the-art performance on two popular action detection benchmarks. On ActivityNet-1.3, it achieves an average mAP of $34.09\%$. On THUMOS14 it reaches $51.6\%$ at IoU@0.5 when combined with a proposal processing method. 




\section{Related Work}
\subsection{Video Representation}
\begin{figure*}
    \centering
    \includegraphics[trim={0.5cm 3cm 1.5cm 2.5cm},width=17cm,clip]{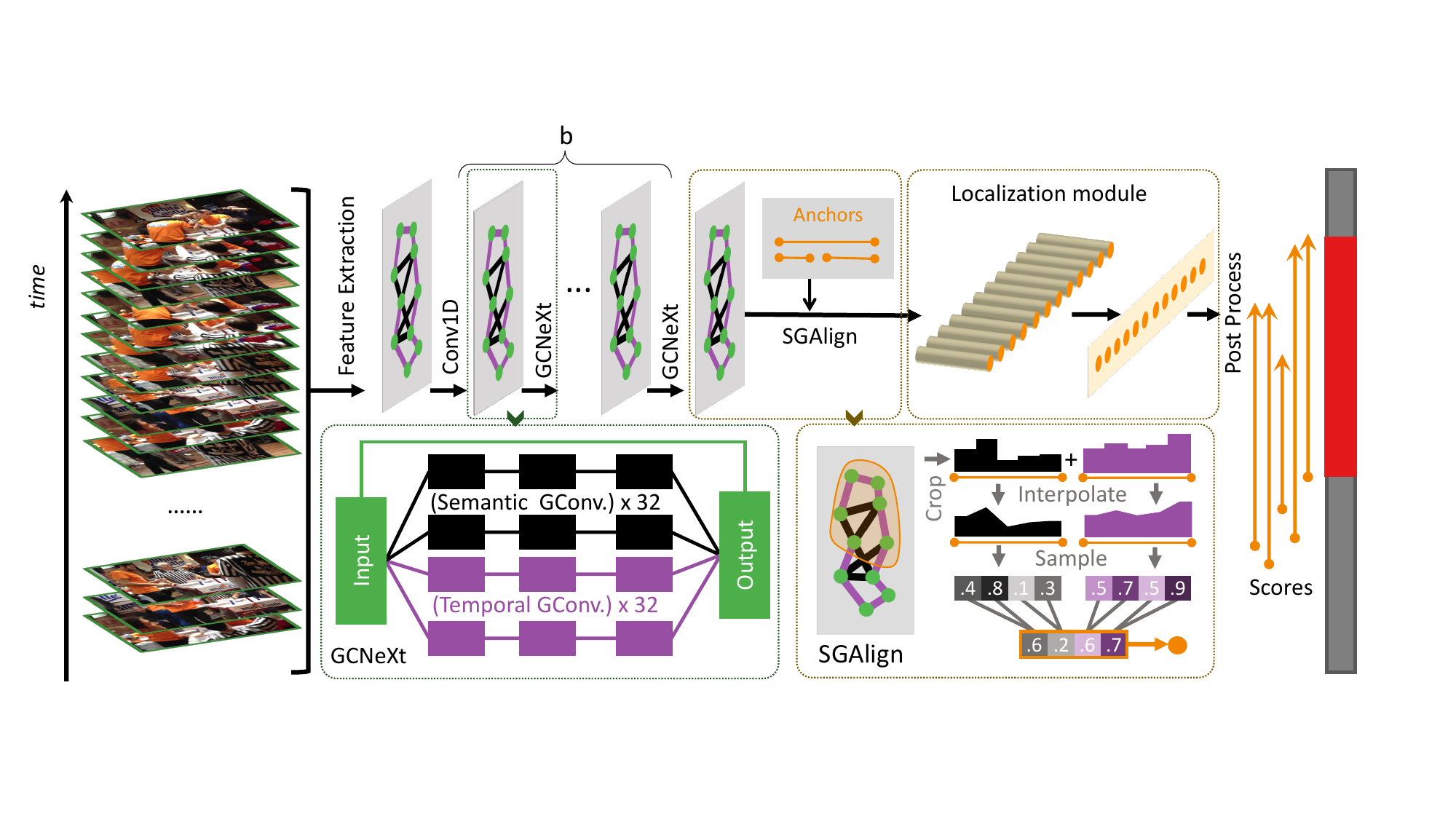}
    \caption{\textbf{Overview of G-TAD architecture.} The input is a sequence of snippet features. We first extract features using $b$ GCNeXt blocks, which gradually aggregate both temporal and multi-level semantic context. Semantic context, encoded in semantic edges, is dynamically learned from features at each GCNeXt layer. 
   Then we feed the extracted features into the  SGAlign layer, where sub-graphs defined by a set of anchors are represented by fixed-size features. 
    Finally, the localization module scores and ranks the sub-graphs for detection. 
    }
    \label{fig:arch}
\end{figure*}

\noindent \textbf{Action Recognition}. Many CNN based methods have been proposed to address the action recognition task. Two-stream networks \cite{feichtenhofer2016convolutional,simonyan2014two,wang2015towards} use 2D CNNs to extract frame features from RGB and optical flow sequences.  
These 2D CNNs can be designed from scratch~\cite{he2016deep, simonyan2014very} or pre-trained on image recognition tasks~\cite{deng2009imagenet}. Other methods~\cite{tran2015learning,carreira2017quo,qiu2017learning,xu2017r} use 3D CNNs to encode spatio-temporal information from the original video. 
In our work, we use the pre-trained action recognition model in \cite{xiong2016cuhk,wang2016temporal} to extract video snippet features as G-TAD input.



\noindent \textbf{Action Detection}. Temporal action detection is to predict the boundaries and categories of action instances  in untrimmed videos. A common practice is to first generate temporal proposals and then classify each proposal into one of the action categories~\cite{shou2016temporal,singh2016untrimmed,zhao2017temporal, zeng2019graph, chao2018rethinking, lin2018bsn}.
For proposal generation, they either use fixed handcrafted anchors \cite{buch2017sst,caba2016fast, escorcia2016daps, gao2017turn,shou2016temporal} , or adaptively form proposal candidates by connecting potential start and end frames
~\cite{zhao2017temporal,lin2018bsn}. 
G-TAD uses anchors to define sub-graphs, but also incorporates start/end prediction to regularize the training process. 

\subsection{GCN in Videos}
\noindent \textbf{Graphs in Video Understanding}. Graphs have been widely used for data/feature representation in various video understanding tasks, such as
action recognition~\cite{liu2019learning, wang2018videos, chen2019graph}, and action localization~\cite{zeng2019graph}.
In action recognition, Liu \textit{et al.} ~\cite{liu2019learning} view a video as a 3D point cloud in the spatial-temporal space.
Wang \textit{et al.} ~\cite{wang2018videos} represent a video as a space-time region graph, in which the graph nodes are defined by object region proposals. In action detection, Zeng \textit{et al.} ~\cite{zeng2019graph} consider temporal action proposals as nodes in a graph, and refine their boundaries and classification scores based on the established proposal-proposal dependencies. Differently from previous works, G-TAD takes video snippets as nodes in a graph and form edges between them based on both their temporal ordering and  semantic similarity. 

\noindent
\textbf{Graph Convolutions Networks.}
Graph Convolutional Networks (GCNs)~\cite{kipf2016semi} are widely used for non-Euclidean structures. In these years, its successful application has been seen in computer vision tasks due to their versatility and effectiveness, such as 3D object detection~\cite{Gkioxari2019Mesh} and point cloud segmentation ~\cite{wang2018dynamic, xie2019clouds}.  
Meanwhile, various GCN architectures are proposed for more effective and flexible modelling. 
One representative work is the edge convolution method by {Wang}
\textit{et al.}~\cite{wang2018dynamic} for point clouds. It computes graph edges (represented as node adjacency) at each graph layer based on the feature distance between nodes, and enriches the node feature by aggregating the features over the neighbourhood as node output.
Recently, Li~\textit{et al.} ~\cite{Li_2019_ICCV,li2019deepgcns_journal} propose DeepGCNs to enable GCNs to go as deep as 100 layers using residual/dense graph connections and dilated graph convolutions, and explore ways to automatically design GCNs~\cite{li2019sgas}.  G-TAD uses a DeepGCN-like structure to apply graph convolutions on a dynamic semantic graph as well as a fixed temporal graph.


\section{Proposed Method}
\subsection{Problem Formulation}
The input to our pipeline is a video sequence of $l_v$  frames.
%
Following recent video action proposal generation methods \cite{buch2017sst, escorcia2016daps, gao2017turn,lin2018bsn}, we construct our G-TAD model using feature sequences extracted from raw video frames. 
We average the features of every $\sigma$ consecutive frames and refer to each set of the $\sigma$ frames as a  \textbf{snippet}.
%
In this way, our input visual feature sequence is represented by
$X^{(0)}\in \mathbb{R}^{C\times L}$, 
where $C$ is the feature dimension of each snippet, and $L$ is the number of snippets. Each video sequence has a set of $N$ annotations
$\Psi  = \left \{ \psi  _n=\left (t_{s,n},t_{e,n}, c_n \right ) \right \}_{n=1}^{N}$, where $\psi_n$ represents an action instance, and $t_{s,n}$, $t_{e,n}$, and $c_n$ are its start time, end time, and action class, respectively.

The temporal action detection task is to predict ${M}$ possible actions $\Phi=  \left \{ \phi _m=\left (\hat{t}_{s,m},\hat{t}_{e,m}, \hat{c}_m, {p}_m \right ) \right \}_{m=1}^{M}$ from  $V$. Here, $(\hat{t}_{s,m},\hat{t}_{e,m})$ represents the predicted temporal boundaries for the $m^{\textrm{th}}$ predicted action; $\hat{c}_m$ and ${p}_m$ are its predicted action class and confidence score, respectively.
%
%

\subsection{G-TAD Architecture}

Our action detection framework is illustrated in Fig.~\ref{fig:arch}. 
We feed the snippet features $X^{(0)}$, into a stack of $b$ GCNeXt blocks, which is designed inspired by ResNeXt~\cite{xie2017aggregated} to obtain context-aware features. Each GCNeXt contains two graph convolution streams. One stream operates on fixed temporal neighbors, and the other adaptively aggregates semantic context into snippet features. 
Both streams follow the split-transform-merge strategy with multiple convolution paths~\cite{xie2017aggregated} (the number of paths is defined as cardinality) to generate updated graphs, which are aggregated into one graph as the block output.
At the end of all $b$ GCNeXt blocks, we extract a set of sub-graphs based on the pre-defined temporal anchors (see Section \ref{subsec: Details}). 

Then we have the sub-graph of interest alignment layer SGAlign to represent each sub-graph
using a feature vector.
In the end, we use multiple fully connected layers to predict the intersection over union (IoU) of the feature vector representing every sub-graph and the ground truth action instance. 
We provide a detailed description of both GCNeXt and SGAlign 
in Sections \ref{subsec: GCNeXt} and \ref{subsec: SGAlign}, respectively.


\begin{figure}
    \centering
    \includegraphics[trim={0cm 0.5cm 0cm 0.5cm},width=8cm,clip] {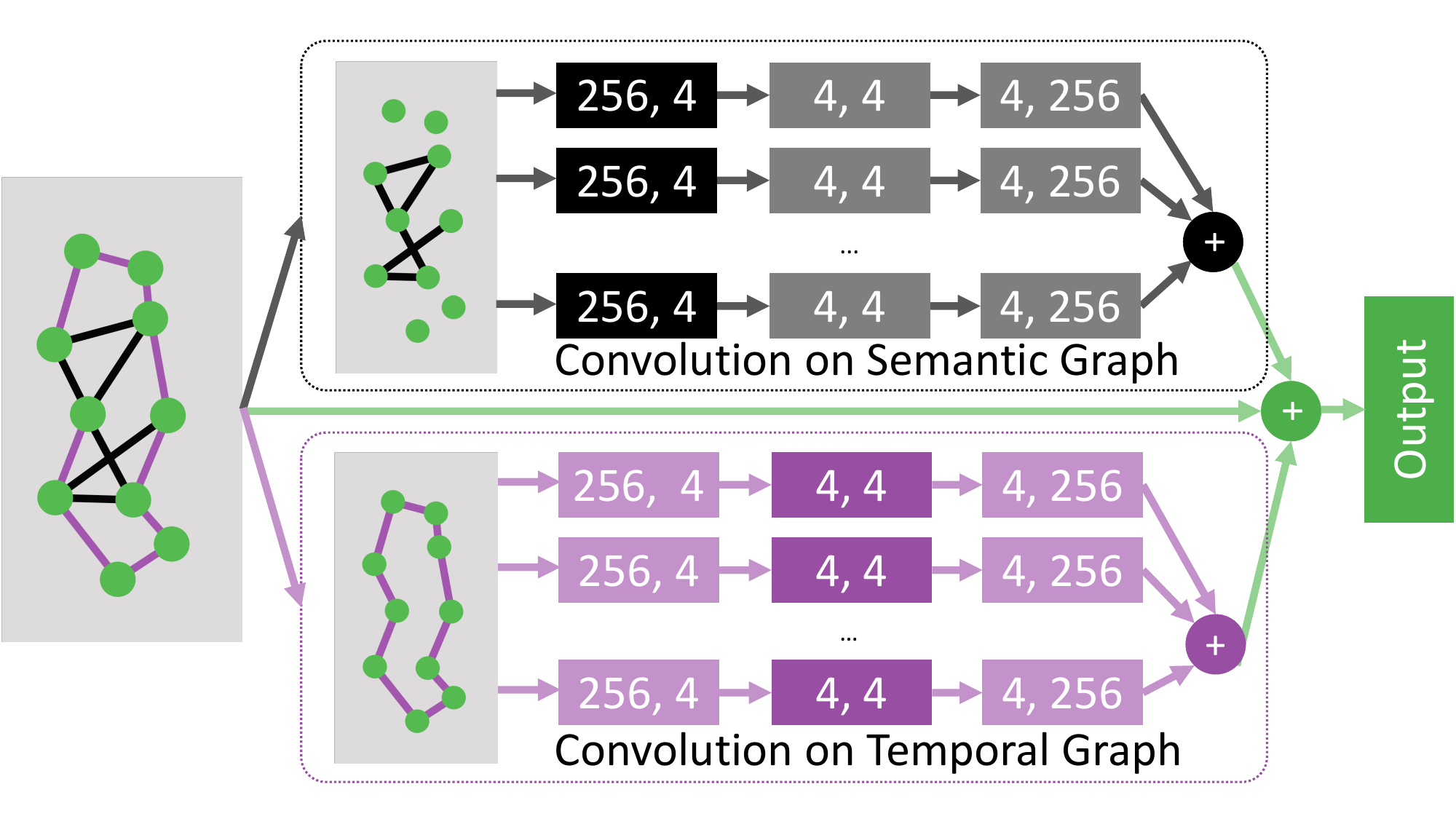}
    \caption{\textbf{GCNeXt block.} The input feature is processed by temporal and semantic streams with the same cardinality. Black and purple boxes represent operations in the temporal and semantic streams, respectively, darker colors referring to graph convolutions and lighter ones 1-by-1 convolutions. The numbers in each box refer to input and output channels. Both streams follow a split-transform-merge strategy with 32 paths each to increase the diversity of transformations. The module output is the summation of both streams and the input.}
    \label{fig:GCNeXt}
\end{figure}

\subsection{GCNeXt for Context Feature Encoding} \label{subsec: GCNeXt}
Our basic graph convolution block, GCNeXt, operates on a graph representation of the video sequence. It encodes snippets using their temporal and semantic neighbors. 
Fig. \ref{fig:GCNeXt} illustrates the architecture of GCNeXt.

We build a video graph $\mathcal{G}=\{\mathcal{V},\mathcal{E}\}$, where $\mathcal{V}=\{v_l\}_{l=0}^L$ and $\mathcal{E}=\mathcal{E}_t \cup \mathcal{E}_s$ denote the node and edge sets, respectively.
In this case, each node represents a snippet and each edge shows a dependency between a pair of snippets. 
We define two types of edges --- temporal edges $\mathcal{E}_t$ and semantic edges $\mathcal{E}_s$. Accordingly we have the temporal stream and the semantic stream. We describe each type of edge as well as the graph convolution process in the following.
%

%


\noindent
\textbf{Temporal Edges ($\mathcal{E}_t$)} encode the temporal order of video snippets. Each node $v_i\in \mathcal{V}$ has one unique forward edge to node $v_{i+1}$, and one backward edge to node  $v_{i-1}$. In this case, we have $\mathcal{E}_t=\mathcal{E}_t^{f}\cup\mathcal{E}_t^{b}$, where  $\mathcal{E}_t^{f}$ and $\mathcal{E}_t^{b}$  are  forward and backward temporal edge sets defined  
as follows:
\begin{align} \label{eq:edge_t}
\mathcal{E}_t^f &=\{(v_i,v_{i+1})|~i\in\{1,2,\dots,L-1\}\}, \\
\mathcal{E}_t^b &=\{(v_i,v_{i-1})|~i\in\{2,\dots,L-1,L\}\},
\end{align}
where \noindent $L$ is the number of snippets in the video.

\noindent \textbf{Semantic Edges ($\mathcal{E}_s$)} are defined from the notion of dynamic edge convolutions \cite{wang2018dynamic}, which dynamically constructs edges between graph nodes according to their feature distances. The goal of our semantic edges is
to collect information from semantically correlated snippets. 
we define the semantic edge set $\mathcal{E}_s$ for each node $v_{i}$ in  $\mathcal{G}$ as follows
\begin{align} 
\mathcal{E}_s=\{(v_i,v_{n_i(k)})|i\in\{1,2,\dots,L\}; k\in\{1,2,\dots K\}\} \notag.
\end{align}
Here, ${n_i(k)}$ refers to the node index of the $k^{\textrm{th}}$ nearest neighbor of node $v_i$. 
It is determined dynamically at every GCNeXt block in the node feature space, enabling us to update the nodes
that intrinsically carry semantic context information throughout the network. Therefore, $\E_s$ adaptively changes to represent new levels of semantic context. 




\noindent \textbf{Graph Convolution}. 
%
We use 
$X=[x_1,x_2,\dots,x_L]\in \mathbb{R}^{C \times L}$ to represent the features for all the nodes in the graph and transform it using the graph convolution operation $\F$. There are several choices for $\F$ in the literature. For simplicity,
we use a single-layer edge convolution~\cite{wang2018dynamic} as our graph convolution operation: 
\begin{align} \label{eq:agg}
   \F(X,A,W)&=([X^T,AX^T-X^T] W)^T .
\end{align}
%
Here, $W\in \mathbb{R}^{2 C \times C'}$ is trainable weight;
$A\in \mathbb{R}^{L \times L}$ is the adjacency matrix without self-loops (\ie edges between a node and itself); $[\cdot,\cdot]$ represents the matrix concatenation of columns.
We formulate the $(i,j)^\textrm{th}$ element in $A$ as $A_{(i,j)} = {\mathbf{1}}\{(v_i,v_j)\in \E\}$, 
where $\mathbf{1}\{\cdot\}$ is the indicator function.
%



%

Either stream in GCNeXt leverages a split-transform-merge strategy~\cite{xie2017aggregated} with 32 paths to increase the diversity of transformations. Each path contains one graph convolution as in Eq.~\ref{eq:agg}  and two 1-by-1 convolutions, their composition denoted as $\F'$.

\noindent \textbf{Stream Aggregation}. 
The GCNeXt output is the aggregation of semantic and temporal steams as well as the input,
which can be formulated as: 
\begin{align} \label{eq:block}
   \H(X,A,W)=ReLU(&\F'{(X,A_t^f,W_t^f)}+\F'{(X,A_t^b,W_t^b)}\nonumber \\
   &+ \F'{(X,A_s,W_s)}+X ),
\end{align}
where $A_t^f$, $A_t^b$, and $A_s$ are adjacency matrices, $W = \{W_t^f, W_t^b, W_s\}$ are the trainable weights, corresponding to  $\E_t^f$, $\E_t^b$, and $\E_s$, respectively. $ReLU$ is the rectified linear unit as the activation function. 
In the \textbf{supplementary material}, we simplify  Eq.~\ref{eq:block}  and prove that it can be efficiently computed by zero-padded 1D convolutions.


\subsection{Sub-Graph Alignment and Localization} \label{subsec: SGAlign}

\begin{figure}
    \centering
    \includegraphics[trim={2.5cm 3.5cm 2.5cm 4.5cm}, width=8cm, clip]{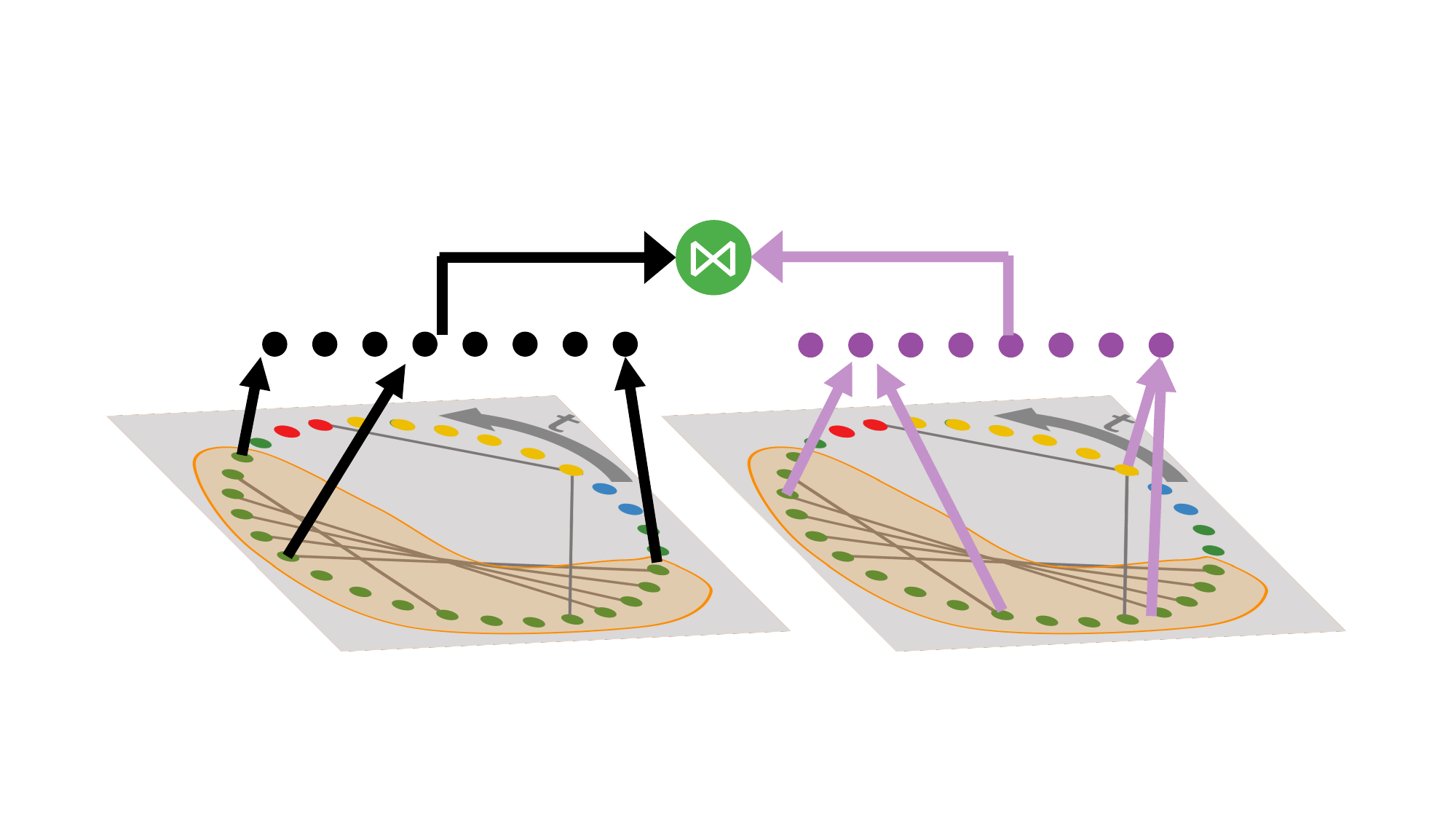}
    \caption{\textbf{SGAlign layer}. SGAlign extracts sub-graph features based on both GCNeXt features (left) and semantic features (right), and concatenates both sub-graph features as output. The dots on top represent sub-graph features. On the bottom, the dots represent graph nodes, grey lines are  semantic edges, and the orange highlighted zones are sub-graphs. Note that since the semantic feature of each node is computed using its neighbors, each entry in the the sub-graph feature essentially corresponds to multiple semantically correlated nodes in the graph.
    }
    \label{fig:ga}
\end{figure}

\noindent
\textbf{Sub-Graph of Interest Alignment (SGAlign)}. 
The GCNeXt blocks generate the features of all snippets $\{x_l\}_{l=1}^L$ (dubbed as GCNeXt features), which contains aggregated information from their temporal and semantic context. Using $\{x_l\}_{l=1}^L$, we obtain an updated graph $\{\mathcal{V}, \mathcal{E}\}$. In SGAlign, we further exploit semantic context  
by averaging the neighbor features of each node, formulated as $y_l = \frac1K\sum_{k=1}^Kx_{n_l(k)}$, and dub $y_l$  as semantic features.

SGAlign uses pre-defined anchors to extract sub-graphs from $\{\mathcal{V}, \mathcal{E}\}$. Given each action anchor $a=(t_s,t_e)$, a sub-graph $\mathcal{G}_a$ is defined as a subset of $\mathcal{G}$ such that  $\mathcal{G}_a=\{\mathcal{V}_a,\mathcal{E}_a\}$, where $\mathcal{V}_a=\{v_l \in \mathcal{V}\} | {t_s\le l\le t_e}\} $ and $\mathcal{E}_a = \{(v_i, v_j) \in \mathcal{E}_s | v_i \in \mathcal{V}_a \}$.
For the sub-graph $\mathcal{G}_a$,  we sample  $\tau$ points ($\tau$: alignment quantity) via interpolation and rescaling as described in Alg. \ref{alg}, and generate the sub-graph feature  $y_a\in \mathbb{R}^{\tau C}$, where $C$ is the feature dimension.

We run Alg. \ref{alg} independently using the GCNeXt features $\{x_l\}_{l=1}^L$ and the semantic features $\{y_l\}_{l=1}^L$ as input. For the former, we sample $\tau_1$ points and obtain the sub-graph features $z_{1a}\in \mathbb{R}^{\tau_1 C}$ ; and for the latter, we sample  $\tau_2$ points and obtain $z_{2a}\in \mathbb{R}^{\tau_2 C}$, respectively. We concatenate $z_{1a}$ and $z_{2a}$ as the output of the SGAlign layer.  Fig.~\ref{fig:ga} illustrates the idea of SGAlign using the two features.

%

\begin{algorithm}[t]

\caption{ Interpolation and Rescaling in SGAlign }
\begin{algorithmic}[1]
\REQUIRE   Features of all nodes in the entire graph $\{x_l\}_{l=1}^L$; sub-graphs $\{\mathcal{G}_{a_j}\}_{j=1}^{J}$, where $J$ is the total number of sub-graphs, $a_j=(t_{s,j}, t_{e,j})$; alignment quantity $\tau$;
\FOR {each sub-graph $\mathcal{G}_{a_j}$}
\STATE Arrange all nodes in $\mathcal{G}_{a_j}$ in their temporal order;
\STATE Compute sub-graph size $d=t_{s,j}-t_{e,j}$, sampling interval $s=\ceil{d/\tau}$,  interpolation quantity  $T$ = $\tau s$;
\STATE Sample $T$ points based on linear interpolation using the two neighbors of each point ${l}$ = [$t_s +  k d/ T $ for $k$ in range($T$)]
\STATE $X_{\textrm{in}}$ = [$(\ceil{i}-i) x_{\floor{i}}+(i-\floor{i}) x_{\ceil{i}}$ for $i$ in ${l}$]
\STATE $z_{a_j}$ = [$\textrm{mean}(X_{in}[k s$:$(k+1)s])$ for $k$ in range($\tau$)]
\ENDFOR
\ENSURE $Z = \{z_{a_j}\}_{j=1}^{J}$.
\end{algorithmic}

\label{alg}
\end{algorithm}

By explicitly using the semantic feature $y_l$, SGAlign  adaptively aggregates semantic context information when computing the features of each sub-graph. This is essentially different from the methods that manually extend the anchor boundaries for incorporating temporal context \cite{lin2018bsn, zhao2017temporal} and leads to superior performance. 

It is worth mentioning that the sampling interval $s$ is based on the sub-graph size $d$ and alignment quantity $\tau$, to ensure that the output $z_{a_j}$ is the weighted sum of \emph{all the nodes} in the sub-graph. In Sec.~\ref{subsec:abl}, we show that this sampling strategy gives us experimental improvements.

\noindent \textbf{Sub-Graph Localization}. For each sub-graph $\G_{a}$, we calculate its Intersection-over-Union (IoU) with all ground-truth actions $\psi$ in $\Psi$, and denote the maximum IoU $g_c$ as the training target.
%
We apply three fully connected (FC) layers on top of the SGAlign layer for each sub-graph feature.
The last FC layer has two output scores $p_{cls}$  and  $p_{reg}$, which are trained to match $g_c$ using classification and regression losses,  respectively.
                
\subsection{G-TAD Training}

We train G-TAD with the sub-graph localization loss $L_{g}$ and the node classification loss $L_n$, as well as an $\mathcal{L}_2$-norm regularization loss $L_r$  for all trainable parameters $\Theta$:

\begin{equation}
L = L_{g} +L_{n}+  \lambda_2 \cdot L_r,
\end{equation}
where we set $\lambda_2=10^{-4}$. 
The loss $L_{g}$ is used to determine the confidence scores of sub-graphs, 
and the loss $L_n$, classifying each node based on its location relative to an action, can drastically improve the network convergence. 

\noindent \textbf{Sub-Graph Localization Loss}. 
The sub-graph loss $L_{g}$ is defined as follows:
%
\begin{equation}
L_{g}=  {L}_{wce}(p_{cls},\mathbf{1}\{g_c > 0.5\}) + \lambda_1 \cdot {L}_{mse}(p_{reg},g_c),
\end{equation}
%
where ${L}_{mse}$ is the mean square error loss and ${L}_{wce}$ is the weighted cross entropy loss. The weight is computed to balance the positive and negative training samples.  In our experiments, we take the trade-off coefficient $\lambda_1=10$, since the second loss term tends to be smaller than the first. 
%

\noindent \textbf{Node Classification Loss}. Along with the sub-graph localization loss $L_{g}$, we use the loss $L_n$ to classify each node in the whole graph based on whether they are start or end points of an action. We add FC layers after the first GCNeXt block to produce the start/end probabilities $(p_s,p_e)$ (these layers are ignored at test time). 
We use $(g_{ns}, g_{ne})$ to denote the corresponding training targets for each node.
%
%
We use the weighted cross entropy loss to compute the discrepancy between the predictions and the targets, and hence have $L_n$ formulated as $L_{n}= {L}_{wce}({{p}}_s,{g}_{ns})+ {L}_{wce}({{p}}_e,{g}_{ne})$. 




\subsection{G-TAD Inference}
At inference time, G-TAD predicts classification and regression scores for each sub-graph $\G_{a}$. From the $J$ sub-graphs, we construct predicted actions $\Phi  = \left \{  \phi_j=(\hat{t}_{s,j}, \hat{t}_{e,j}, \hat{c}_j, p_j) \right \}_{j=1}^{J}$, where $(\hat{t}_{s,j}, \hat{t}_{e,j})$ refer to the predicted action boundaries, $\hat{c}_j$ is the predicted action class, and $p_j$ is the fused confidence score of this prediction, computed as $p_{j}=p_{cls}^\alpha \cdot p_{reg}^{1-\alpha}$. In our experiments, we search for the optimal $\alpha$ in each setup. 
\section{Experiment}
\providecommand{\e}[1]{\ensuremath{\times 10^{#1}}}
\subsection{Datasets and Metrics}


\noindent
{\bf ActivityNet-1.3} \cite{caba2015activitynet}  is a large-scale action understanding dataset for action recognition, temporal detection, proposal generation and dense captioning tasks. It contains 19,994 temporally annotated untrimmed videos with 200 action categories, which are divided into training, validation and testing sets by the ratio of 2:1:1. 

\noindent
{\bf THUMOS-14} \cite{jiang2014thumos} dataset contains 413 temporally annotated untrimmed videos with 20 action categories. We use the 200 videos in the validation set for training and evaluate on the 213 videos in the testing set. 

\noindent
{\bf Detection Metric}. We take mean Average Precision (mAP) at certain IoU thresholds as the main evaluation metric. 
Following the official evaluation API, the IoU thresholds $\{0.5, 0.75, 0.95\}$ and $\{0.3, 0.4, 0.5, 0.6, 0.7\}$ are used for ActivityNet-1.3 and THUMOS-14, respectively. On ActivityNet-1.3,  we also report average mAP over 10 different IoU thresholds $[0.5: 0.05:0.95]$.

\subsection{Implementation Details} \label{subsec: Details}

\noindent
{\bf Features and Anchors}. We use pre-extracted features for both datasets. For ActivityNet-1.3, we adopt the pre-trained two-stream network by Xiong et. al.~\cite{xiong2016cuhk}, with down-sampling ratio $\sigma = 16$. 
Each video feature sequence is rescaled to $L = 100$ snippets using linear interpolation.
For THUMOS-14, the video features are extracted using TSN model~\cite{TSN2016ECCV} pre-trained on Kinetics~\cite{zisserman2017kinetics}  with $\sigma = 5$.
We crop each video feature sequence with overlapped windows of size $L=256$ and stride $128$. In training, we do not use any crops void of actions.

For ActivityNet-1.3 and THUMOS-14, we enumerate all possible combinations of start and end as anchors, \eg $\{(t_s,t_e)|~ 0< t_s<t_e<L; ~t_s,t_e \in \mathcal{N}; ~t_e-t_s < D\} $, where $D$ is $100$ for ActivityNet-1.3 and $64$ for THUMOS-14. 
In SGAlign, we use $\tau_1=32,\tau_2=4$ for ActivityNet-1.3 and $\tau_1=\tau_2=16$  for THUMOS-14. 


\noindent {\bf Training and Inference}. We implement and compile our framework using PyTorch 1.1, Python 3.7, and CUDA 10.0.
We use $b=3$ 
GCNeXt blocks and train our model end-to-end, with batch size of 16. 
The learning rate is $4\e{-3}$ on ActivityNet-1.3 and 
$6\e{-6} $ on THUMOS-14 for the first 5 epochs, and is reduced by 10 for the following 5 epochs.
%
In inference, following \cite{lin2018bsn}  to leverage the global video context, we take the video classification scores from \cite{wang2017untrimmed} and \cite{xiong2016cuhk}, and multiply them by the fused confidence score $p_j$ for evaluation.
For post-processing, we apply Soft-NMS \cite{softNMS}, where the threshld is $0.84$ and select the top-$M$ prediction for final evaluation, where $M$ is 100 for  ActivityNet-1.3 and 200 for THUMOS-14.
More details can be found in the \textbf{supplementary material}.
 

\subsection{Comparison with State-of-the-Art}

\noindent
{\bf ActivityNet-1.3}: Tab.~\ref{tab:sota_anet} compares G-TAD with state-of-the-art detectors. We report mAP at different tIoU thresholds, as well as average mAP. G-TAD reports the highest average mAP results on this large-scale and diverse dataset. Notably, G-TAD reaches an mAP of 9.02\% at IoU 0.95, indicating that the localization is more accurate than others.


\begin{table}[tbp]
\centering
\caption{\textbf{Action detection results on validation set of ActivityNet-1.3}, measured by mAP ($\%$) at different tIoU thresholds and the average mAP. G-TAD achieves better performance in average mAP than the other methods, even the latest work of BMN and P-GCN shown in the  second-to-last block. 
}
\small
\begin{tabular}{p{2.1cm}p{0.62cm}<{\centering}p{0.62cm}<{\centering}p{0.62cm}<{\centering}p{0.9cm}<{\centering}}
\toprule
Method  & 0.5  &  0.75  & 0.95 & Average\\
\hline
Wang \textit{et al.} \cite{wang2016uts}    & 43.65 & -  & - & -\\
Singh \textit{et al.} \cite{singh2016untrimmed} & 34.47 & - & - & - \\
SCC \cite{heilbron2017scc}   & 40.00 & 17.90  & 4.70   & 21.70  \\
CDC \cite{shou2017cdc} & 45.30 & 26.00 & 0.20 & 23.80 \\
R-C3D \cite{xu2017r} & 26.80 & - & - & - \\
BSN \cite{lin2018bsn} & 46.45  & 29.96 & 8.02  & 30.03  \\
Chao \textit{et al.} \cite{chao2018rethinking} & 38.23 & 18.30 & 1.30 & 20.22 \\ \hline
P-GCN \cite {zeng2019graph} &48.26 &33.16 &3.27 &31.11  \\
BMN \cite{lin2019bmn} & { 50.07} & \textbf{34.78} & { 8.29} & { 33.85}  \\
\hline
\textbf{G-TAD} (ours)  & \textbf{50.36} & {34.60} & \textbf{9.02} & {\bf 34.09} \\
\bottomrule
\end{tabular}
\label{tab:sota_anet}
\end{table}

\begin{table}[!tb]
\centering
\caption{\textbf{Action detection results on testing set of THUMOS-14}, measured by mAP (\%) at different tIoU thresholds. G-TAD achieves the best performance for IoU@0.7, and combined with P-GCN, G-TAD significantly outperforms all the other methods. 
}
\small
\begin{tabular}{p{2.5cm}p{0.7cm}<{\centering}p{0.7cm}<{\centering}p{0.7cm}<{\centering}p{0.7cm}<{\centering}p{0.7cm}<{\centering}}
\toprule
    Method                                         &  0.3 &  0.4 &  0.5 &  0.6 &  0.7 \\ \hline
    SST \cite{buch2017sst}                 & - & -    & 23.0 & - & -\\
    CDC  \cite{shou2017cdc}             & 40.1 & 29.4 & 23.3 & 13.1 & 7.9\\
    TURN-TAP\cite{gao2017turn}                 & 44.1 & 34.9 & 25.6 & - & -\\
    CBR  \cite{gao2017cascaded}      & 50.1 & 41.3 & 31.0 & 19.1 & 9.9\\
    SSN \cite{zhao2017temporal}             & 51.9 & 41.0 & 29.8 & - & -\\ 
    BSN \cite{lin2018bsn}                 & 53.5 & 45.0 & 36.9 & 28.4 & 20.0 \\
    TCN \cite{dai2017temporal}            & - & 33.3 & 25.6 & 15.9 & 9.0\\
    TAL-Net \cite{chao2018rethinking}         & 53.2 & {48.5} & {42.8} & {33.8} & 20.8\\
    MGG \cite{DBLP:journals/corr/abs-1811-11524} & 53.9 & 46.8 & 37.4 & 29.5 & 21.3 \\
    DBG \cite{lin2019fast} & {57.8} & 49.4 & 39.8 & 30.2 & {21.7} \\
    Yeung \textit{et al.} \cite{yeung2016end}             & 36.0 & 26.4 & 17.1  & - & - \\
    Yuan \textit{et al.} \cite{Yuan2017}                    & 36.5 & 27.8 & 17.8 & - & -\\
    Hou \textit{et al.} \cite{hou2017real}                 & 43.7 & -    & 22.0 & - & -\\
    SS-TAD  \cite{buch2017end}                & 45.7 & - & 29.2 & - & 9.6\\
    BMN \cite{lin2019bmn}                  & {56.0} & 47.4 & 38.8 & 29.7 & 20.5 \\
    {G-TAD} (ours)                                 & {54.5} & {47.6} & {40.2} & {30.8} & \textbf{23.4}  \\
    \hline
    BSN+P-GCN~\cite{zeng2019graph}                 & {63.6} & {57.8} & {49.1} & - & -\\
    \textbf{G-TAD}+P-GCN                                & \textbf{66.4} & \textbf{60.4} & \textbf{51.6} & \textbf{37.6} & {22.9}  \\
\bottomrule
\end{tabular}
\label{Tab:sota_thm}
\end{table}

\noindent
{\bf THUMOS-14}: Tab.~\ref{Tab:sota_thm} compares the action localization results of G-TAD and various state-of-the-art methods on the THUMOS14 dataset. At IoU 0.7, G-TAD reaches an mAP of
23.4\%, obviously higher than the current best 20.8\% of TALNet. At IoU 0.5, G-TAD outperforms all methods except TALNet.
Besides, when combined with a proposal post-processing method P-GCN~\cite{zeng2019graph}, G-TAD performs even better, especially at IoUs $\leq 0.5$. Now G-TAD reaches 51.6\% at IoU 0.5, outperforming all the other methods. In addition, we also report the results of BSN with P-GCN (directly taken from~\cite{zeng2019graph}), which are not as good as G-TAD + P-GCN, albeit showing improvement from BSN. This signifies the advantage of G-TAD proposals regardless of post-processing.

\subsection{Ablation Study} \label{subsec:abl}

\noindent
\textbf{GCNeXt Module}: We ablate the three main components of GCNeXt, mainly GCN on temporal edges, GCN on semantic edges, and cardinality increase. 
%
Tab.~\ref{tab:abl_gcnext} reports the performance on ActivityNet-1.3, where each component is separately enabled/disabled. We see how each of these components contributes to the performance of the final G-TAD model. We highlight the gains from the semantic graph, showing the benefit of integrating adaptive context from semantic neighbors. It also shows cardinality 32 mostly outperforms cardinality 1.

\begin{table}[tbp]
\centering
\caption{\textbf{Ablating GCNeXt Components.} We disable temporal/semantic graph convolutions and set different cardinalities for detection on ActivityNet-1.3.   
}
\small
\begin{tabular}{p{0.7cm}<{\centering}p{0.7cm}<{\centering}p{0.7cm}<{\centering}|p{0.65cm}<{\centering}p{0.65cm}<{\centering}p{0.65cm}<{\centering}p{0.55cm}<{\centering}p{0.65cm}<{\centering}}
\hline
\multicolumn{3}{c|}{ GCNeXt block} & \multicolumn{4}{c}{tIoU on Validation Set}\\
Temp. & Sem.  & Card.  & 0.5  &  0.75  & 0.95  & Avg. \\
\hline
\xmark & \xmark & 1 & {48.12} & {32.16} & 6.41 & {31.65} \\ \hline
\cmark & \cmark & 1 & {50.20} & \textbf{34.80} & 7.35 & {33.88} \\
\cmark & \xmark & 32& {50.13} & {34.17} & {8.70} & {33.67} \\
\xmark & \cmark & 32& {49.09} & {33.32} & 8.02 & {32.63} \\ \hline
\cmark & \cmark & 32& {\textbf{50.36}} & {{34.60}} & \textbf{9.02} & {\textbf{34.09}} \\
\hline
\end{tabular}
\label{tab:abl_gcnext}
\end{table}

\noindent \textbf{SGAlign Module}: 
The incorporation of semantic features in SGAlign aggregates more semantic context into each sub-graph, which benefits the subsequent localization compared to merely using the GCNeXt features. The sampling interval $s$ in Alg.~\ref{alg} is adaptively computed for each sub-graph, leading to better performance than a fixed value (\textit{e.g.} $s=1$).
Tab.~\ref{tab:abl_sg} shows the effect of  semantic features and the sampling strategy from both temporal and semantic graphs on ActivityNet-1.3. While sampling densely gives us minor improvements, we obtain a larger gain by including context information from the semantic graph.

\begin{table}[tbp]
\centering
\caption{\textbf{Ablating SGAlign Components.
} 
We disable the sample-rescale process and the feature concatenation from the semnantic graph for detection on ActivityNet-1.3. The rescaling strategy leads to slight improvement, while the main gain arises from the use of context information (semantic graph).  }
\small
\begin{tabular}{p{1.0cm}<{\centering}p{1.0cm}<{\centering}|p{0.65cm}<{\centering}p{0.65cm}<{\centering}p{0.65cm}<{\centering}p{0.45cm}<{\centering}p{0.65cm}<{\centering}}
\hline
\multicolumn{2}{p{2cm}<{\centering}|}{ SGAlign} & \multicolumn{4}{c}{tIoU on Validation Set}\\ 
Samp. & Concat.  &  0.5  &  0.75  & 0.95  & Avg. \\
\hline
\xmark & \xmark     & {49.84} & {34.58} & 8.17 & {33.78} \\ \hline
\cmark & \xmark     & {49.86} & \textbf{{34.60}} & \bf{9.56} & {33.89} \\
\cmark & \cmark  & \bf{50.36} & \textbf{{34.60}} & 9.02 & \bf{34.09} \\
\hline
\end{tabular}
\label{tab:abl_sg}
\end{table}

\noindent\textbf{Sensitivity to Video Length}: We report the results of the sensitivity of G-TAD to different window sizes in THUMOS-14 in Tab.~\ref{tab:abl_window}. G-TAD benefits more from larger window sizes ($L=256$ vs. $128$).
Larger windows mean that G-TAD can aggregate more context information from the semantic graph. Performance degrades at $L=512$,
where GPU memory limits the batch size and network training is influenced.

\begin{table}[!tb]
\centering
\caption{\textbf{Effect of Video Size.} We vary the input video size (window length $L$) and see that G-TAD performance improves with larger sizes ($L=256$). Degradation occurs at $L=512$, since GPU memory limits the batch size to be significantly reduced, leading to a noticeable  performance drop.}
\small
\begin{tabular}{p{1.5cm}<{\centering}|p{0.6cm}<{\centering}p{0.6cm}<{\centering}p{0.6cm}<{\centering}p{0.6cm}<{\centering}p{0.6cm}<{\centering}}
\hline
Window & \multicolumn{5}{c}{tIoU on Validation}  \\
    Length $L$    &  0.3 &  0.4 &  0.5 &  0.6 &  0.7 \\ \hline
    64              &47.07&39.29& 32.25 &23.88& 15.17\\
    128             & 51.75 & 44.90 & 38.70 & 29.03 & 21.32\\
    256             & \textbf{54.50} & \textbf{47.61} & \textbf{40.16} & \textbf{30.83} & \textbf{23.42} \\
    512             & 48.32 & 41.71 & 34.38 & 26.85 & 19.29 \\ \hline
\end{tabular}
\label{tab:abl_window}
\end{table}

\begin{figure}
    \centering
    \includegraphics[trim={3cm 2cm 3cm 2cm},width=8cm,clip]{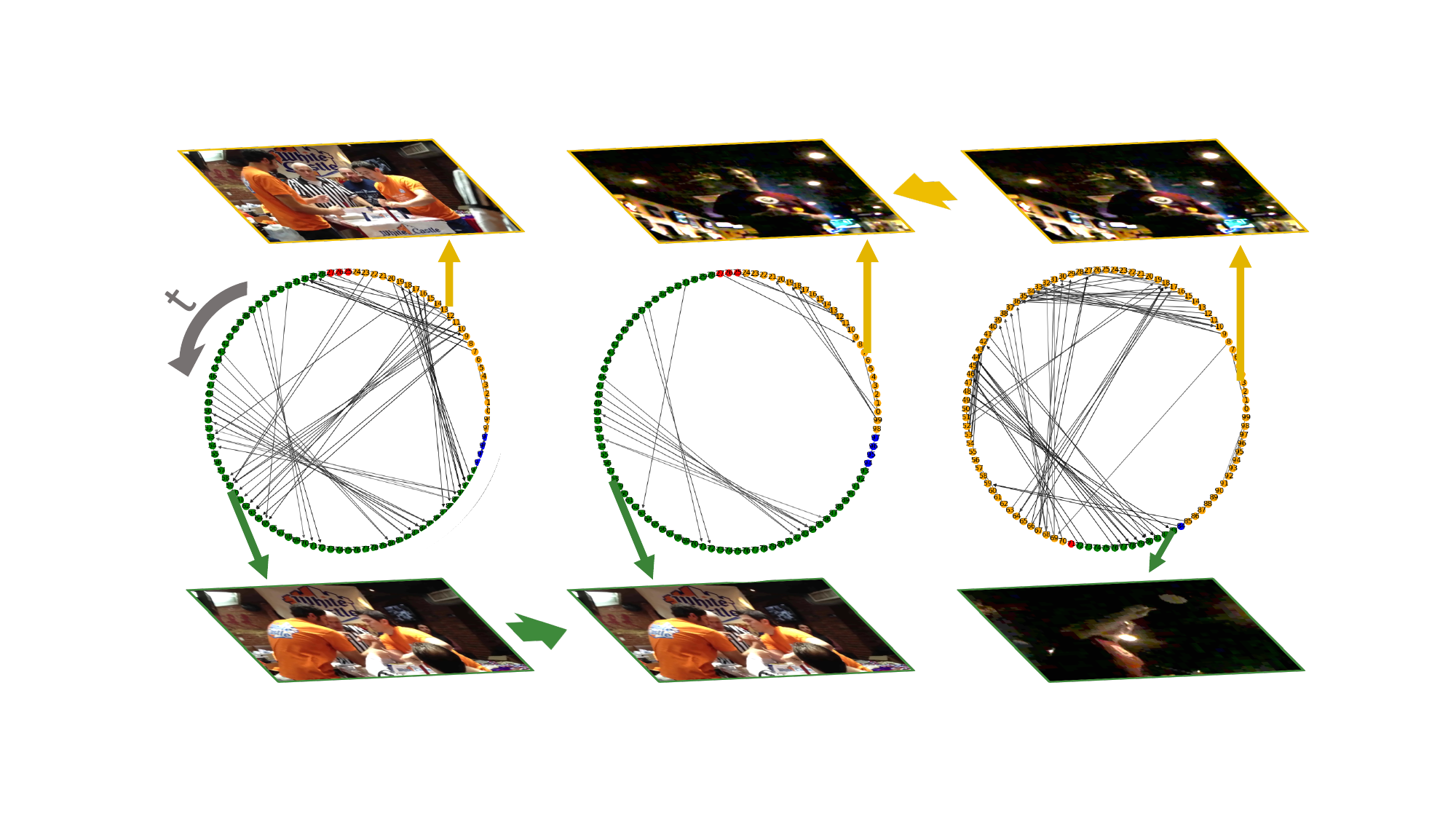}
    \caption{\textbf{Semantic graphs and Context}. Given two 
    videos (left and right), we combine action frames of one video with background frames of another to create a synthetic video with no action context (middle). As expected, the semantic graph of the synthetic video contains no edges between action and background snippets. 
    } 
    \label{fig:qual_results}
\end{figure}

\begin{figure*}[!h]
    \centering
    \vspace{-0.5cm}
    \includegraphics[trim={0cm 4cm 0cm 4cm}, width=14cm,clip]{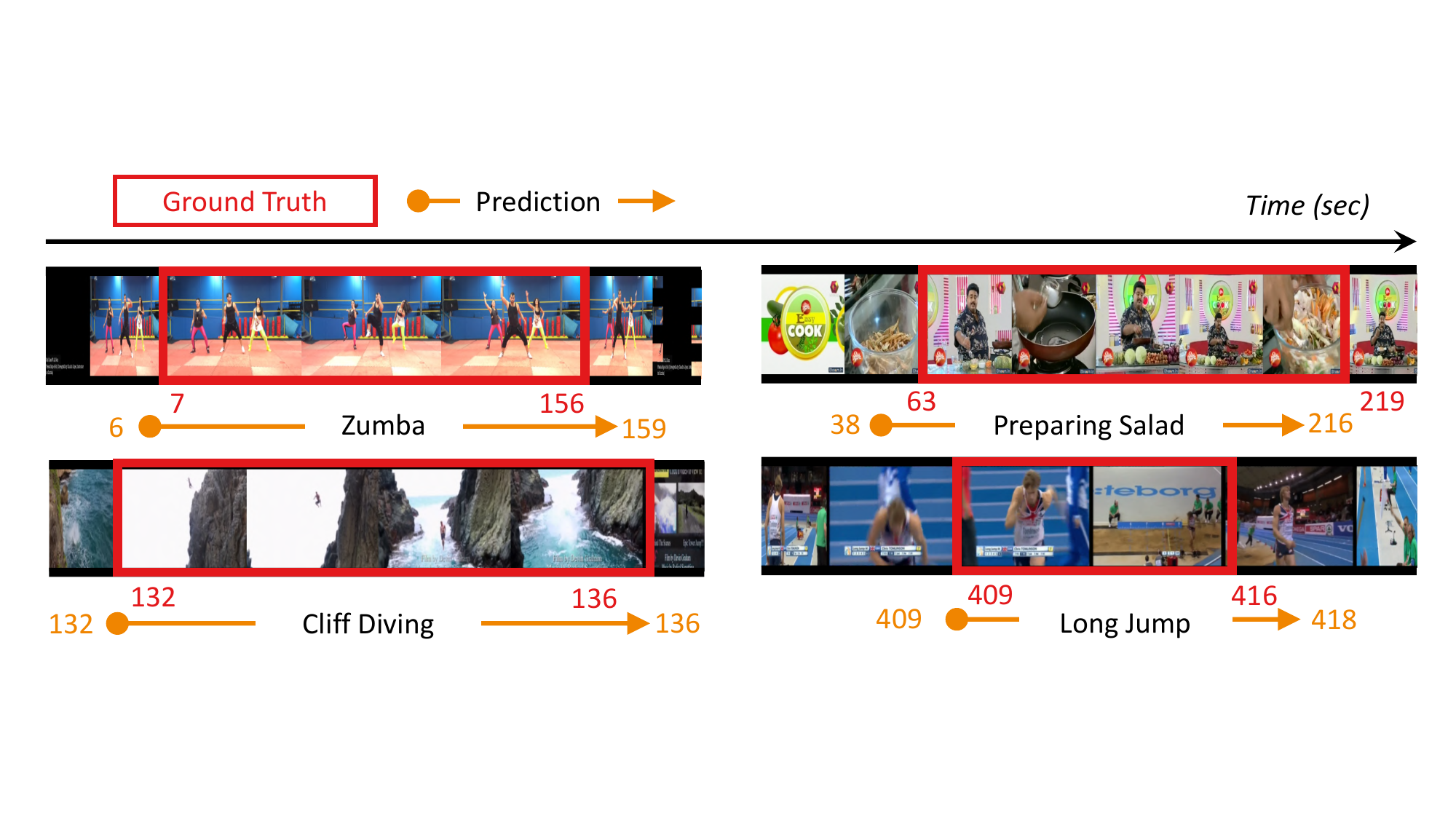}
    \caption{\textbf{Qualitative results.} We show qualitative detection results on ActivityNet-1.3 (top) and THUMOS-14 (bottom).}
    \label{fig:vis_video_THUMOS}
\end{figure*}

\subsection{Discussion of Action Context} 

In the ablation study, graph convolutions on the semantic graph improve G-TAD performance in both the GCNeXt block and in the SGAlign layer. Semantic edges connecting background to action snippets can adaptively pass the action context information to each possible action. In this section, we define 2 extra experiments to show how semantic edges encode meaningful context information.

\noindent {\bf Zero-Context Video}. How zero context between action and background leads to semantic graphs with no action-background edges is visually shown by comparing semantic graphs resulting from natural videos and synthetically compiled ones.
In Fig.~\ref{fig:qual_results} (left and right), we present two natural videos that  include actions ``wrestling" and ``playing darts", respectively. Semantic edges in their resultant graphs do exist, connecting action with background snippets, thus exemplifying the usage of context in the detection process.
%
Then, we compile a synthetic video that stacks action frames from the wrestling video and background frames from the darts video, feed it to G-TAD and again visualize the semantic graph (middle). As expected, the semantic graph does not include any action-background semantic edges.

\begin{figure}[!ht]
    \centering
    \includegraphics[trim={0cm 1cm 0cm 1cm},width=7cm,clip]{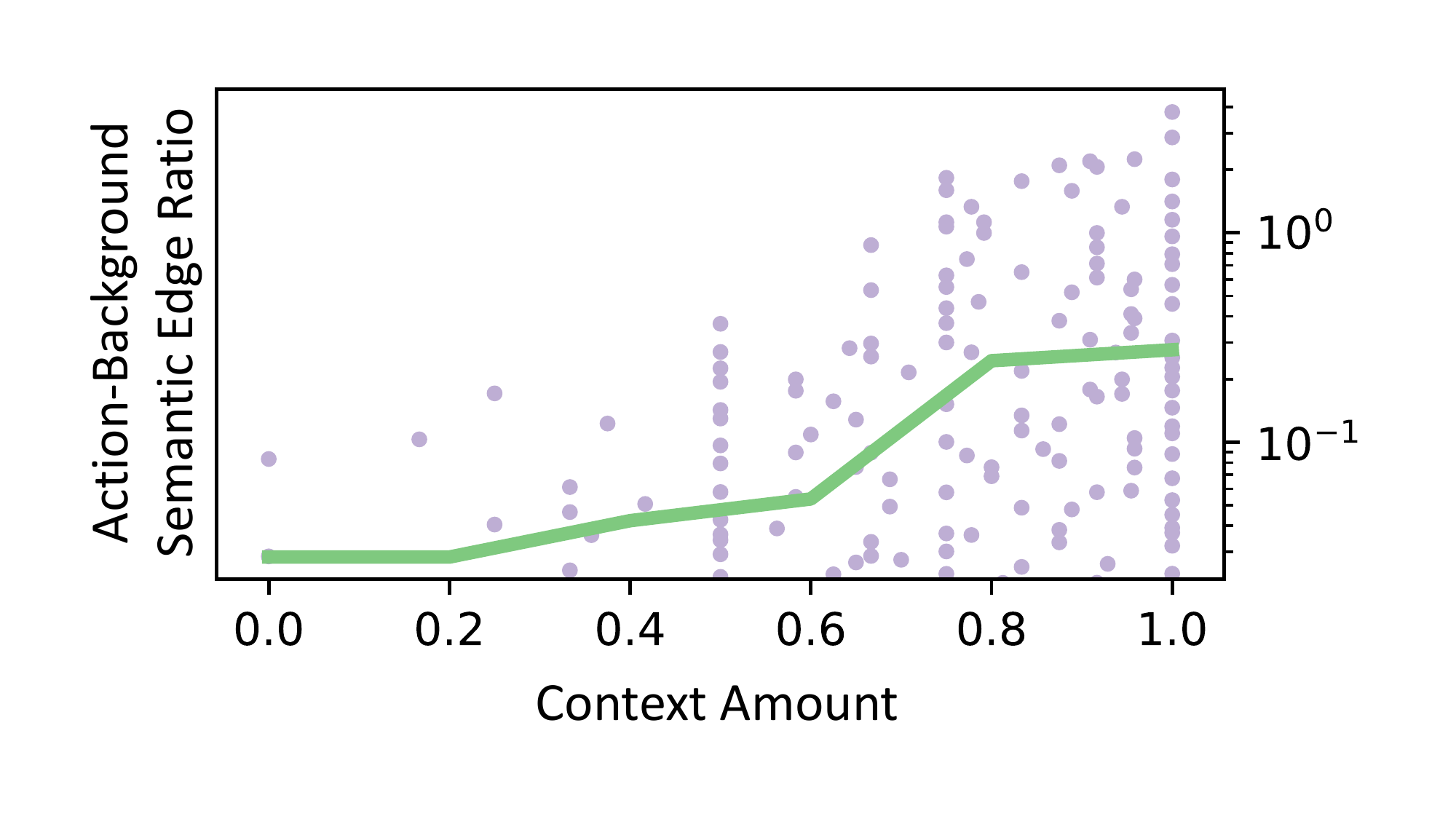}
    \caption{\textbf{Action-Background Semantic Edge Ratio vs. Context Amount.} In the scatter plot, each purple dot corresponds to a different video graph. 
     Strong positive correlation is observed between context amount and action-background  semantic  edge ratio, which means we predict on average more semantic edges in the presence of larger video context.
    }
    \label{fig:ratio}
\end{figure}

\begin{figure}[!ht]
    \centering
    \includegraphics[trim={1.5cm 0.1cm 1cm 0cm},width=7.5cm,clip]{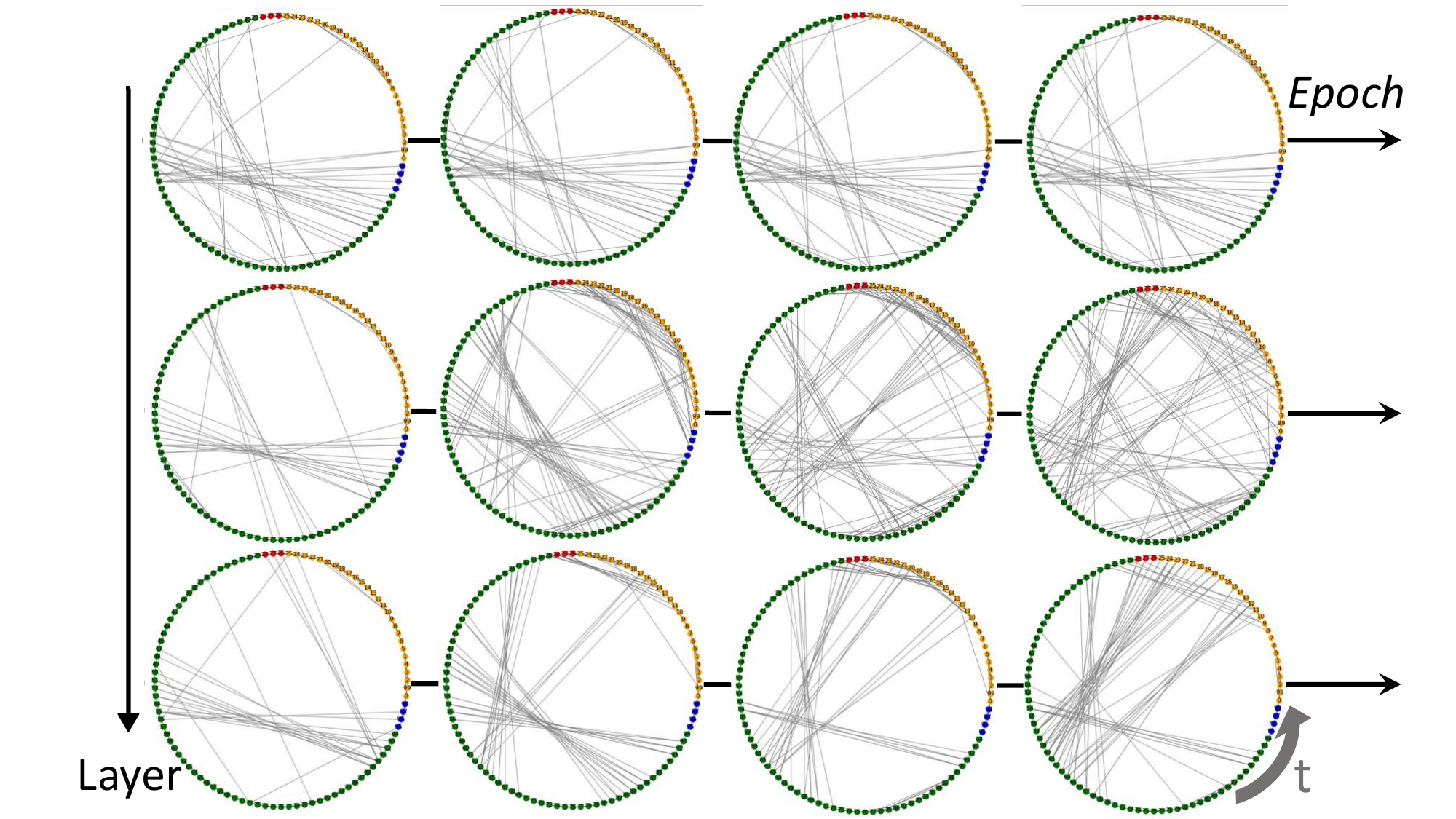}
    \caption{\textbf{Semantic graph evolution during G-TAD training.} We visualize the semantic graphs at first, middle, and last layers during training epoch 0, 3, 6, and 9. The semantic edges at the first layer are always the same, while the semantic graphs at the middle and last layers evolve to incorporate more context.}\vspace{-6pt}
    \label{fig:evol}
\end{figure}

\noindent{\bf Correlation to Context Amount}. We show the correlation between context edges and context as defined by human annotators. 
We define the video \textbf{context amount} as the average number of background snippets which can be used to predict the foreground class. 
Following DETAD~\cite{detad}, we collect context amount for all videos in ActivityNet validation set from Amazon Mechanical Turk. 
The scatters in Fig.~\ref{fig:ratio} shows the relation between \textbf{context amount} and 
the ratio of \textbf{action-background} semantic edges over all the semantic edges.
From the plot, we observe that if a video has a higher amount of context, it is more likely to have more action-background semantic edges in its semantic graph. We further average the ratios in five context amount ranges, and plot them in green. The strong positive correlation between {context amount} and {action-background semantic edge ratio} indicates that our G-TAD model can effectively find related context snippets in the semantic graph.

\subsection{Visualization}
We show a few qualitative detection results in Fig.~\ref{fig:vis_video_THUMOS} on both ActivityNet-1.3 and THUMOS-14. In Fig.~\ref{fig:evol}, we visualize the evolution of semantic graphs during the training process across GCNeXt layers. Specifically, we feed a video into G-TAD and visualize the semantic graphs emerging at the first, middle, and last layers at epochs 0, 3, 6, and 9 of training. The semantic graphs at the first layer are the same, since they are built on the same input features. As we progress to different layers and epochs, semantic graphs adaptively update their edges. Interestingly, we observe the presence of more context edges as training advances. This indicates that G-TAD progressively learns to incorporate multiple levels of context in the detection process. 

\section{Conclusion}
In this paper, 
we cast the temporal action detection task as a sub-graph localization problem by formulating videos as graphs.
We take video snippets as graph nodes, snippet-snippet correlations as edges, and apply graph convolution as the basic operation.
We propose a new architecture G-TAD to localize sub-graphs. G-TAD includes GCNeXt blocks to aggregate context features from semantically correlated snippets  and an SGAlign layer to transform sub-graph features into vector representations. 
G-TAD can learn enriched multi-level semantic context in an adaptive way using stacked dynamic graph convolutions.
Extensive experiments show that G-TAD can find global video context without extra supervision and achieve the state-of-the-art performance on both THUMOS-14 and ActivityNet-1.3.

\noindent
{\textbf{Acknowledgments.} This work was supported by the King Abdullah University of Science and Technology (KAUST) Office of Sponsored Research (OSR) under Award No. OSR-CRG2017-3405.}

\newpage
{\small
\bibliographystyle{ieee_fullname}
\bibliography{egbib}
}

\onecolumn
\clearpage
\newpage
\section{Supplementary}
\setcounter{section}{0}
\renewcommand{\thesection}{\Alph{section}}
\renewcommand{\theHsection}{appendixsection.\Alph{section}}

\section{Derivation and Efficient Implementation of Eq.~{\color{red} 4}}

In this section, we provide the derivation of Eq.~{\color{red} 4} in the paper (listed here in the following). We also show that Eq.~{\color{red} 4} can be  efficiently implemented by zero-padded 1D/edge convolutions. \newline

\begin{align} \label{eq:block2}
   ~~~~~~~~~~~~~~~~~~~~~~~~~~~~~~~~~~~~~~~~~~\H(X,\E,W)=ReLU[&W_\alpha X + W_\beta X A_t^f + W_\gamma X A_t^b+ _\phi X A_s+X]. ~~~~~~~~~~~~~~~~~~~~~~~~~~~~~~~~~~~~~~~ \textrm{(4)}\nonumber
\end{align}

\noindent
\subsection{Derivation of Eq. 4}

\noindent
\textbf{a) Temporal Graph Convolution.} We first provide the derivation for temporal graph convolution.

The temporal forward edges $\mathcal{E}_t^f$ and backward edges $\mathcal{E}_t^b$ are formulated as
\begin{align} 
    \mathcal{E}_t^f=\{(v_i,v_{i+1})|~i\in\{1,2,\dots,L-1\}\} , \mathcal{E}_t^b=\{(v_i,v_{i-1})|~i\in\{2,\dots,L-1,L\},
\end{align}

The corresponding adjacency matrices $A_t^f, A_t^b$ can be present by $L$ vectors, respectively, shown in Eq.~\ref{eq:tgc}. We use $e_k\in \mathbb{R}^{L}$ to present the vector in which the $k$-th element is one but the others are zeros. 
\begin{align} \label{eq:tgc}
    A_t^f = &
        \begin{bmatrix}
        0 & 0 & 0 & \dots & 0 & 0 \\
        1 & 0 & 0 &   & 0 & 0 \\
        0 & 1 & 0 &  & 0 & 0 \\
        \vdots&  &  & \ddots & \vdots & \vdots \\
         0 & 0 & 0 & \dots & 0 & 0 \\
        0 & 0 & 0 & \dots & 1 & 0 \\
        \end{bmatrix} 
    = [e_2,e_3,\dots, e_L, 0] \nonumber\\
    A_t^b = &
        \begin{bmatrix}
        0 & 1 & 0 & \dots& 0 & 0 \\
        0 & 0 & 1 &   & 0 & 0 \\
        0 & 0 & 0 &  & 0 & 0 \\
         \vdots&  &  & \ddots & \vdots & \vdots \\
          0 & 0 & 0 & \dots & 0 & 1 \\
        0 & 0 & 0 & \dots & 0 & 0 \\
        \end{bmatrix}    
    = [0, e_1, e_2,\dots, e_{L-1}], \nonumber\\
    \Rightarrow & A_t^f = [A_t^b]^T
\end{align}

Given the input $X\in\mathbb{R}^{C\times L}$, after temporal graph convolution, the output, $X_t$, becomes
\begin{align} \label{eq:tgc}
    X_t &= f_{agg}(X,A_t^f,W_f) + f_{agg}(X,A_t^b,W_b) =W_{f,0} X + W_{f,1} X A_t^f + W_{b,0} X + W_{b,1} X A_t^b \nonumber\\
    & =(W_{f,0}+W_{b,0}) X + W_{f,1} X A_t^f + W_{b,1} X A_t^b .
\end{align}
Here $W_*$ is the trainable weights in the neural network.

\noindent
\textbf{b) Semantic Graph Convolution.}  It is straightforward to obtain Eq.~{\color{red} 4} for semantic graph convolution.

\noindent
\subsection{Efficient Implementation of Eq.~{\color{red} 4}}

In implementation of Eq.~{\color{red} 4}, we use an efficient zero-padded 1D convolution and edge convolution for temporal graph convolution and semantic graph convolution, respectively. In the following, we provide proof that our efficient implementation is equivalent to Eq.~{\color{red} 4}.

\noindent
\textbf{a) Temporal Graph Convolution.}

If a 1D convolution has kernel size 3, the weight matrix is a 3D tensor in $\mathbb{R}^{3\times C \times C}$. We denote the matrix as $W_{conv1}= [W_1,W_2,W_3],W_{1,2,3}\in \mathbb{R}^{C\times C}$. 
Given the same input $X=[x_1, x_2,...,x_L]$, we pad zero on the input, $x_0=x_{L+1}=0\in \mathbb{R}^{C}$. 

The output of 1D convolution can be written as $Y=[y_1,y_2,...,y_L]\in \mathbb{R}^{C\times L}$  
\begin{align} \label{eq:y_k}
    y_k = [W_1 x_{k-1}+W_2 x_k+W_3x_{k+1}], k=1,2,\dots,L
\end{align}

We can prove that $X_t=Y$ by multiplying $e_k$ on both sides in Eq.~\ref{eq:proof}.  Please be noted that $W_*$ is the trainable weights in the neural network. We can assume $W_1=W_{b,1},W_3=W_{f,1}, W_2=W_{f,0}+W_{b,0}$
\begin{align} \label{eq:proof}
    X_te_k&=W_{f,0} Xe_k + W_{f,1} X A_t^fe_k + W_{b,0} Xe_k + W_{b,1} X A_t^be_k \nonumber\\
    &=(W_{f,0}+W_{b,0}) x_k + W_{f,1} X [0, e_1, e_2,\dots, e_L]^T e_k + W_{b,1} X [e_2,e_3,\dots, e_L, 0]^T e_k  \nonumber\\
    &=(W_{f,0}+W_{b,0}) x_k + W_{f,1} X e_{k+1} + W_{b,1} X e_{k-1} \nonumber\\
    &=(W_{f,0}+W_{b,0}) x_k + W_{f,1} x_{k+1} + W_{b,1} x_{k-1}  \nonumber\\
    &=W_1 x_{k-1}+W_2 x_k+W_3X_{k+1}
\end{align}

\noindent
\textbf{b) Semantic Graph Convolution.} 
In the semantic graph, edge convolution is directly used, so proof is done.

\newpage
\section{Training Details}

\noindent
\textbf{Semantic Edges from Multiple Levels.}
In G-TAD, we use multiple GCNeXt blocks to adaptively incorporate multi-level semantic context into video features. After that, SGAlign layer embeds each sub-graph by concatenating aligned features from temporal and semantic graphs. However, it is not necessary to consider \textbf{only the last} GCNeXt semantic graphs to align the semantic feature. Last row in Tab.~\ref{tab:abl_sgoi} present one more experiment that takes \textbf{the union of} semantic edges from all GCNeXt blocks to aggregate the semantic feature. We can find that the semantic context also helps to improve model performance under this setup.
\begin{table}[h]
\centering
\caption{\textbf{Ablating SGAlign Components.
} 
We disable the sample-rescale process and the feature concatenation from the semnantic graph for detection on ActivityNet-1.3. The rescaling strategy leads to slight improvement, while the main gain arises from the use of context information (semantic graph).  }
\small
\begin{tabular}{p{1.0cm}<{\centering}p{1.0cm}<{\centering}|p{0.65cm}<{\centering}p{0.65cm}<{\centering}p{0.65cm}<{\centering}p{0.45cm}<{\centering}p{0.65cm}<{\centering}}
\hline
\multicolumn{2}{p{2cm}<{\centering}|}{ SGAlign} & \multicolumn{4}{c}{tIoU on Validation Set}\\ 
Samp. & Concat.  &  0.5  &  0.75  & 0.95  & Avg. \\
\hline
\xmark & \xmark     & {49.84} & {34.58} & 8.17 & {33.78} \\ \hline
\cmark & \xmark     & {49.86} & {{34.60}} & \bf{9.56} & {33.89} \\
\cmark & \cmark  & \bf{50.36} & {{34.60}} & 9.02 & \bf{34.09} \\ \hline
\cmark & all  & {50.26} & \bf{34.70} & 8.52 & {33.95} \\
\hline
\end{tabular}
\label{tab:abl_sgoi}
\end{table}


\noindent
\textbf{2D Conv. for Sub-Graph Localization.} 
Once we get the \textit{sub-graph feature} from SGAlign layer, instead of using three fully connected (FC) layers regress to $g_c$, we can arrange the anchors in a 2D $L\times L$ map based on the start/end time, and set zeros to the map where is no pre-designed anchors (e.g. $t_s>t_e$). In doing so, we can use 2D CNNs to regress to a $g_c$ map that arranged by the same order. We call the predicted matrix \textbf{IoU map}.

The neighbouring anchors in the 2D IoU map have similar boundary locations. Thus we can use the proposal-proposal relationship in the 2D convolutions. We set kernel size to 1, 3, and 5, and the results are shown in Tab.~\ref{tab:abl_Conv_on_IOU_map}. We do not observe any significant benefit from 2D convolutions.

\begin{table}[h]
\centering
\caption{\textbf{The model performance when we use 3 2D convolution layers to predict IoU map.}We set kernel size to 1, 3, and 5, and collect result on ActivityNet1.3. We do not observe any significant benefit from 2D convolutions.}
\small
\begin{tabular}{p{1.5cm}<{\centering}p{0.7cm}<{\centering}|p{0.65cm}<{\centering}p{0.65cm}<{\centering}p{0.65cm}<{\centering}p{0.55cm}<{\centering}p{0.65cm}<{\centering}}
\hline
\multicolumn{2}{p{3.2cm}|}{ Conv. on IoU map} & \multicolumn{4}{c}{mAP on Validation Set}\\
Kernel Size & Padding  &  0.5  &  0.75  & 0.95  & Avg. \\
\hline
(1,1) & (0,0)   & {\textbf{50.25}} & {34.66} & \textbf{9.29} & {34.08} \\
(3,3) & (1,1)   & {\textbf{50.25}} & {\textbf{34.94}} & 7.74 & {\textbf{34.10}} \\
(5,5) & (2,2)   & {49.88} & {34.39} & 8.96 & {33.77} \\
\hline
\end{tabular}
\label{tab:abl_Conv_on_IOU_map}
\end{table}

\end{document}